\documentclass[10pt,twocolumn,letterpaper]{article}
\usepackage{graphicx}
\usepackage{amsmath,amssymb,amsfonts}
\usepackage{booktabs}
\usepackage{multirow}
\usepackage[table]{xcolor}
\usepackage{algorithm}
\usepackage{algorithmic}
\usepackage{subcaption}
\usepackage{bm}
\usepackage{nicefrac}

\usepackage[pagenumbers]{cvpr}

\usepackage{subcaption}
\usepackage{pifont}
\usepackage{microtype}
\usepackage{bm}
\newcommand{\vecb}[1]{\bm{#1}}

\definecolor{cvprblue}{rgb}{0.21,0.49,0.74}
\usepackage[pagebackref,breaklinks,colorlinks,allcolors=cvprblue]{hyperref}
\makeatletter
\def\abstract{%
   \iftoggle{cvprpagenumbers}{}{%
     \thispagestyle{empty}%
   }%
   \centerline{\large\bf Abstract}%
   \vspace*{3pt}%
   \noindent\itshape\ignorespaces%
}

\makeatother

\title{Disentangled Anatomy-Disease Diffusion (DADD) for Controllable Ulcerative Colitis Progression Synthesis}

\author{
Umut Dundar, Alptekin Temizel\\
Graduate School of Informatics\\
Middle East Technical University, Turkey\\
{\tt\small dundar.umut@metu.edu.tr, atemizel@metu.edu.tr}
}

\begin{document}
\maketitle
\begin{abstract}
Synthesizing longitudinal medical images at controllable disease stages while preserving patient-specific anatomy is hindered by the entanglement of pathological textures and structural features. We address this challenge for ulcerative colitis (UC) endoscopy, where severity follows a continuous ordinal progression along the Mayo Endoscopic Score (MES). Our framework, Disentangled Anatomy-Disease Diffusion (DADD), conditions a latent diffusion model on two complementary embeddings: a pretrained image encoder for patient anatomy and a separately trained ordinal embedder for cumulative disease severity. Since image embeddings inevitably capture disease information, we introduce a Feature Purifier, a cross-attention-based erasure mechanism that identifies and suppresses disease-correlated channels, yielding purified anatomical representations. 
These cleaned anatomy tokens and target disease tokens are injected into the denoising network via a Triple-Pathway Cross-Attention mechanism with resolution-dependent routing gates. This architecture leverages the U-Net hierarchy, in which different network depths encode global structure versus fine-grained pathological texture. Furthermore, we introduce Delta Steering, a training-free directional signal derived from the ordinal embeddings that enables explicit, single-pass control over disease transitions at inference without requiring additional forward passes. Validated on the LIMUC dataset, our approach produces high-fidelity images across all severity levels and effectively rebalances skewed class distributions, enhancing performance for downstream classification tasks. The dataset is available at \href{https://zenodo.org/records/5827695}{zenodo.org/records/5827695} and the code base at \href{https://github.com/umutdundar99/progressive-stable-diffusion}{github.com/umutdundar99/progressive-stable-diffusion}
\end{abstract}

\section{Introduction}
\label{sec:introduction}

\begin{figure}[t]
  \centering
  \includegraphics[width=0.9\linewidth]{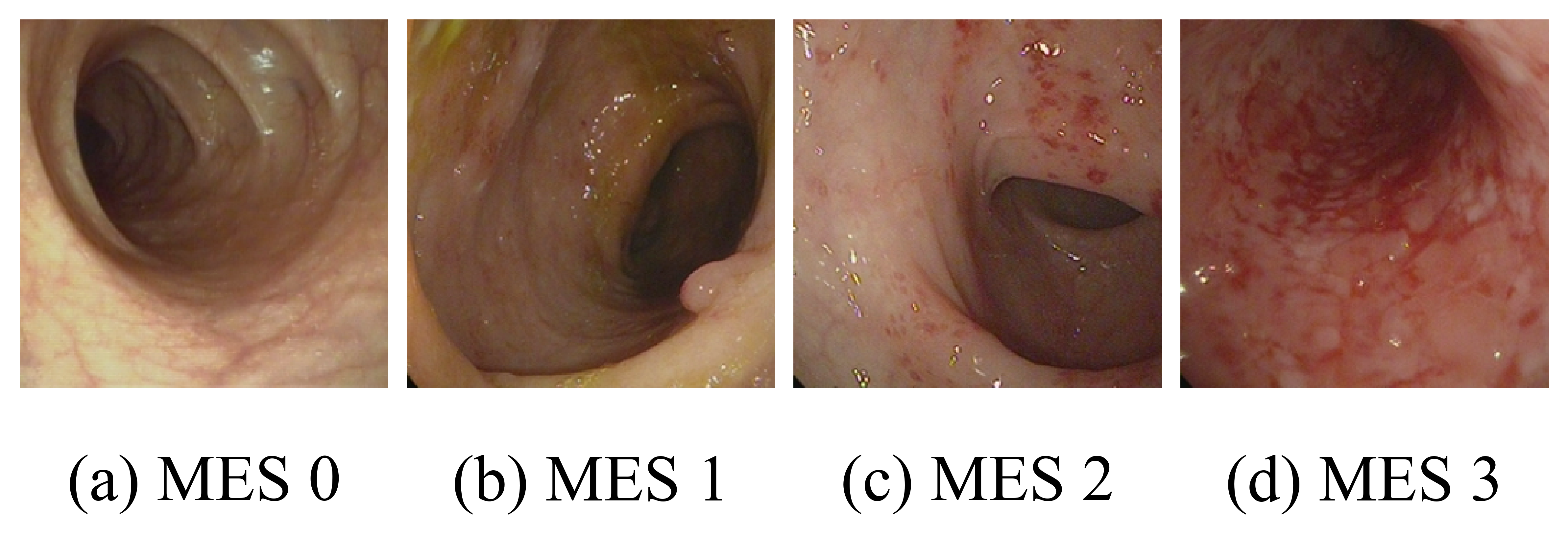}
  \vspace{-3mm}
  
  \caption{\textbf{Sample labeled images from LIMUC~\cite{polat2022labeled}.} 
  \textbf{(a)} normal mucosa with visible vascular pattern. 
  \textbf{(b)} mild erythema, decreased vascular pattern. 
  \textbf{(c)} marked erythema, friability, erosions. 
  \textbf{(d)} spontaneous bleeding and ulceration.}
  \label{fig:mayo_ref}
\end{figure}

Generative models for image synthesis are advancing rapidly, with significant impact in the medical field, where data scarcity and privacy constraints often limit access to large-scale datasets. Traditionally, Generative Adversarial Networks (GANs)~\cite{frid2018gan, yi2019generative} have been widely adopted to synthesize medical images and augment training sets for downstream clinical tasks. However, recent focus has shifted towards Diffusion Models~\cite{kazerouni2023diffusion}, which offer superior training stability, better mode coverage, and higher image quality. A key advantage of modern diffusion architectures is their highly adaptable U-Net backbones, which readily integrate diverse conditioning signals (e.g., text prompts, clinical labels, or reference images~\cite{rombach2022high, ye2023ip}), enabling precise, controllable generation. A compelling clinical application of these models is the synthesis of gastrointestinal disease images, such as ulcerative colitis (UC). UC severity is assessed using the Mayo Endoscopic Score (MES)~\cite{schroeder1987coated}, an ordinal scale ranging from 0 (normal) to~3 (severe) (\cref{fig:mayo_ref}). Existing datasets, such as the Labeled Images for Ulcerative Colitis (LIMUC) dataset~\cite{polat2022labeled}, provide valuable resources but suffer from severe class imbalance and lack longitudinal follow-up due to the invasive nature of repeated endoscopies. Synthesizing subject-specific longitudinal data computationally would enable visualization of disease progression and provide targeted augmentation for downstream classification tasks.

\begin{figure*}[t]
  \centering
  \includegraphics[width=0.75\linewidth]{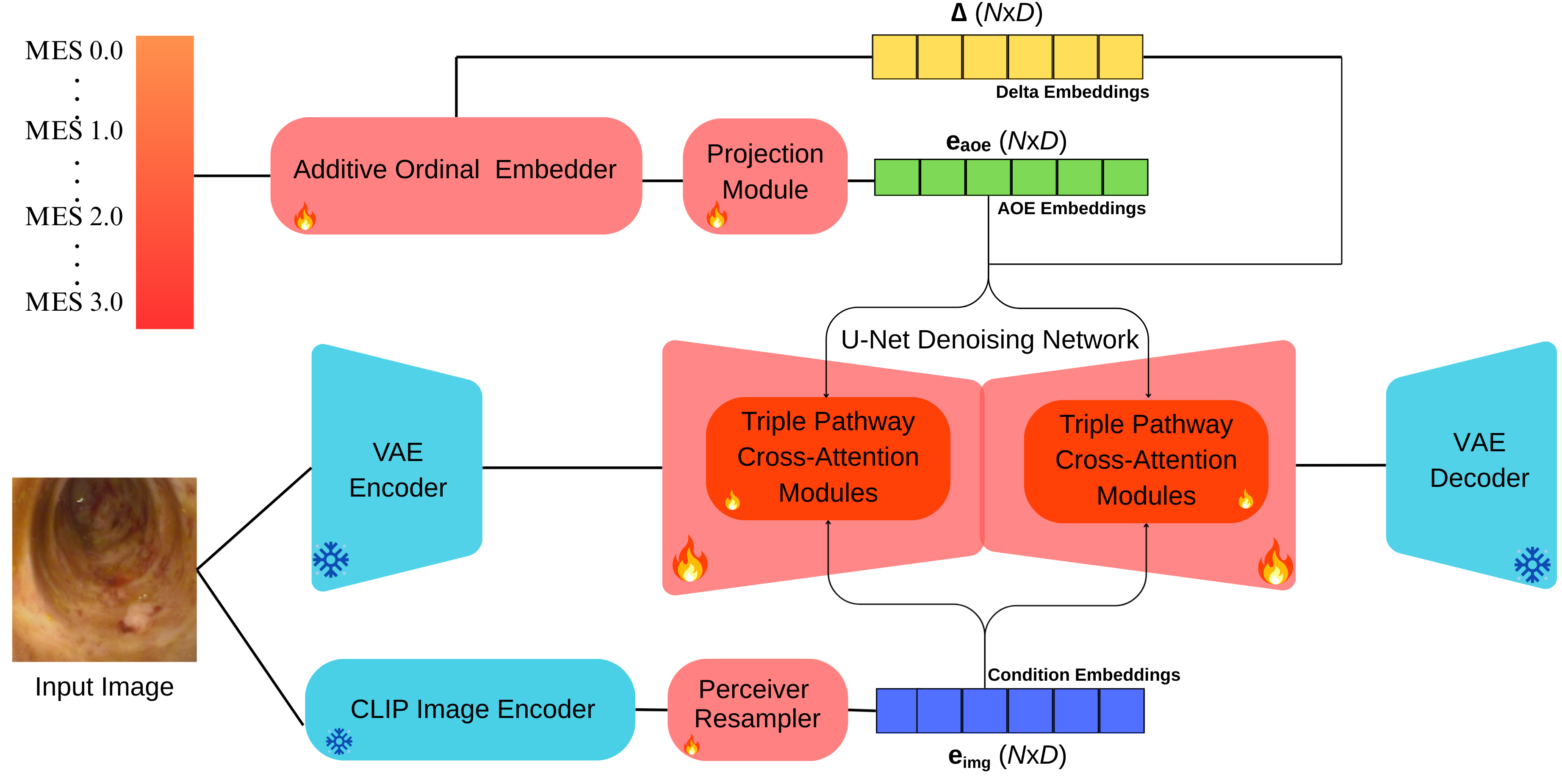}
  \caption{\textbf{Framework overview.} The input image $\mathbf{I}$ is mapped into the latent space by the frozen encoder of the Variational Autoencoder (VAE)~\cite{rombach2022high}. Simultaneously, a frozen CLIP image encoder (ViT-L/14)~\cite{radford2021learning} extracts feature tokens from $\mathbf{I}$, which are then compressed into $N$ anatomy tokens via a trainable Perceiver Resampler~\cite{jaegle2021perceiver}. The AOE~\cite{kurt2025progressive} and Projection Module encode the target MES as a cumulative ordinal embedding into $N$ disease tokens. Delta embeddings are extracted as the post-projection of two MES scores. All three token streams are injected into the U-Net denoising network via Triple Pathway Cross-Attention (as detailed in \cref{fig:attention_detail}). During inference, the VAE's frozen decoder maps the denoised latent representations back to pixel space to generate the final synthetic image. Snowflake and flame icons indicate frozen and trainable components, respectively.}
  \label{fig:pipeline}
\end{figure*}

Recent work effectively models continuous MES progression using an Additive Ordinal Embedder (AOE)~\cite{kurt2025progressive}; however, because it synthesizes samples from random noise, it inherently alters patient-specific anatomy. While IP-Adapter~\cite{ye2023ip} can condition generation on reference images using the Contrastive Language-Image Pretraining (CLIP) image encoder~\cite{radford2021learning}, it inevitably entangles patient-specific anatomy (e.g., mucosal folds) with source disease pathology (e.g., ulceration). This creates a direct semantic conflict when attempting to alter the target severity.

We ground the development and evaluation of our DADD framework in the LIMUC dataset, the current benchmark for labeled UC endoscopic images. Our key insight is to force anatomy and disease embeddings to interact via cross-attention, explicitly erasing disease information from the image representation. By querying image tokens with the AOE, pathological channels are identified and gated out, yielding a purified anatomical representation. We inject these tokens into the U-Net via frequency-aware routing gates, thereby confining disease edits to fine-texture scales while preserving patient-specific structural identity. Finally, we introduce \emph{delta steering}, a single-pass signal derived from the projected AOE difference that provides magnitude-proportional control over changes in severity.

\noindent Our contributions are as follows:
\begin{itemize}[leftmargin=*,nosep]
\item Cross-attention-based disease erasure that purifies image tokens using ordinal embeddings, disentangling anatomy from pathology.
\item Frequency-aware split-injection attention with fixed routing gates across U-Net layers.
\item Delta steering, a single-pass alternative to Classifier-free Guidance (CFG)~\cite{ho2022classifier}, providing signed, magnitude-proportional control of disease changes. This approach reduces inference latency by $\approx 2\times$ compared to standard CFG.

\item Empirical evaluation showing improved generation fidelity and enhanced downstream classification performance via targeted augmentation.
\end{itemize}

\section{Related Work}
\label{sec:related}

Latent diffusion models~\cite{rombach2022high} have largely replaced GANs for high-fidelity synthesis~\cite{dhariwal2021diffusion} and are increasingly adopted in medical imaging, including chest X-rays~\cite{chambon2022roentgen} and histopathology~\cite{aversa2024diffinfinite}. In GI endoscopy, recent work demonstrated that ordinal embeddings in a fine-tuned stable diffusion model can successfully capture ulcerative colitis progression~\cite{kurt2025progressive}. However, while these models generate visually plausible disease states, they struggle to preserve instance-specific anatomical structures across varying severity levels.

To preserve structural identity, a parallel line of work conditions diffusion models on reference images. While methods like ControlNet~\cite{zhang2023adding} and T2I-Adapter~\cite{mou2024t2i} enforce strict spatial alignment using dense representations such as edge maps or depth, they impose heavy constraints on the localized structural deformations needed to model severe pathologies, such as ulceration. Instead, semantic conditioning approaches such as IP-Adapter~\cite{ye2023ip} inject CLIP image features~\cite{radford2021learning} via decoupled cross-attention, and its Plus variant adds a Perceiver Resampler~\cite{jaegle2021perceiver} for richer spatial encoding.

To tackle the entanglement between style and content within these semantic spaces, methods like DEADiff~\cite{qi2024deadiff} and Inv-Adapter~\cite{xing2025inv} employ frequency-band Q-Formers and inverted split-injection, respectively. We adopt a similar decomposition philosophy; however, rather than separating generic artistic styles from natural images, we specifically target the disentanglement of pathological disease texture from patient anatomy in the medical domain.

Concept erasure methods employ diverse mechanisms to remove unwanted attributes: LEACE~\cite{belrose2023leace} utilizes fixed linear projections; recent methods such as SuMa~\cite{nguyen2025suma} and GrOCE~\cite{han2025groce} achieve erasure through target subspace mapping and graph-guided semantic severing, respectively. Unlike these approaches, our Feature Purifier learns a \emph{conditional} gate that dynamically adapts the degree of erasure to the disease severity present in the source image. At the attention level, methods like Prompt-to-Prompt~\cite{hertz2022prompt}, CASteer~\cite{gaintseva2025casteer}, and NASA~\cite{nguyen2025supercharged} steer generation by intervening in the cross-attention mechanism, such as by modifying attention maps or injecting steering vectors into hidden representations. Our delta steering shares this fundamental principle but operates uniquely within an ordinal embedding space specifically designed for monotonic disease progression, offering a single-pass alternative to CFG.

Synthetic augmentation is a promising remedy for endemic class imbalance in medical datasets, traditionally explored with GANs~\cite{shin2018medical} and more recently with diffusion models~\cite{trabucco2023effective}. In contrast to generic generation, our work emphasizes preserving patient-specific anatomy while enabling controlled, monotonic edits of disease severity. This targeted severity control produces longitudinally plausible samples, directly addressing the skewed distributions and subsequently improving downstream classification performance.
\providecommand{\vecb}[1]{\boldsymbol{#1}}
\providecommand{\mat}[1]{\mathbf{#1}}
\providecommand{\eclean}{\vecb{e}_{\mathrm{clean}}}
\providecommand{\eimg}{\vecb{e}_{\mathrm{img}}}
\providecommand{\eaoe}{\vecb{e}_{\mathrm{aoe}}}
\providecommand{\edelta}{\vecb{\Delta}}
\providecommand{\ganat}{g_{\mathrm{a}}}
\providecommand{\gdis}{g_{\mathrm{d}}}
\providecommand{\zanat}{\vecb{z}_{\mathrm{a}}}
\providecommand{\zdis}{\vecb{z}_{\mathrm{d}}}
\providecommand{\zdelta}{\vecb{z}_{\delta}}
\providecommand{\wmod}{\mat{W}_{\mathrm{mod}}}

\section{Method}
\label{sec:method}

We present a conditional latent diffusion framework designed to disentangle instance-specific anatomy from disease severity, enabling controlled, monotonic progression synthesis from a single endoscopic image (\cref{fig:pipeline}).

\subsection{Motivation: The Entanglement Bottleneck}
\label{subsec:problem}

To condition the generation process on both patient anatomy and target disease severity, a naive approach would combine an image prompt with an ordinal severity embedding. Given a source image $\mathbf{I}$ with MES $y_s \in \{0, 1, 2, 3\}$, IP-Adapter~\cite{ye2023ip} extracts condition $\eimg = \mathcal{E}_{img}(\mathbf{I})$ via a pre-trained CLIP image encoder $\mathcal{E}_{img}$ and Perceiver Resampler. Simultaneously, the target severity $y_t$ is encoded via AOE~\cite{kurt2025progressive} into disease tokens $\eaoe^{\mathrm{t}} = \text{AOE}(y_t)$. Ideally, $\eimg$ strictly encodes anatomical identity $A$, while $\eaoe^{\mathrm{t}}$ dictates target pathology. In practice, CLIP embeddings entangle anatomy and source disease $D_s$, yielding $\eimg \approx A \oplus D_{s}$.

When attempting to generate a counterfactual progression where $y_t \neq y_s$, injecting both $\eimg$ and $\eaoe^{\mathrm{t}}$ into the U-Net's cross-attention layers creates a direct semantic conflict. The network receives competing pathological signals: the residual source disease $D_{s}$ from the image tokens and the intended target disease $D_{t}$ from the AOE. This entanglement prevents monotonic disease progression and leads to structural artifacts, as the model attempts to render two distinct severity levels simultaneously. 

\subsection{Proposed Architecture}
\label{subsec:architecture}

To resolve this entanglement, the disease information $D_{s}$ must be explicitly identified and erased from $\eimg$ prior to the denoising process. To this end, our proposed framework comprises three key components:
\begin{enumerate}[label=(\arabic*), leftmargin=*]
\item \textbf{Additive Ordinal Embedder (AOE):} Encodes disease severity ($y_t$) as cumulative pathological tokens $\eaoe^{\mathrm{t}}$ (\cref{subsec:aoe}).
\item \textbf{Feature Purifier:} A cross-attention-based mechanism that removes disease-correlated information ($D_{s}$) from $\eimg$, yielding purified anatomy tokens $\eclean$ (\cref{subsec:purifier}).
\item \textbf{Triple-Pathway Cross-Attention \& Delta Steering:} A split-injection attention mechanism with resolution-aware routing gates, coupled with delta steering for inference-time severity control.
(\cref{subsec:routing,subsec:delta}).
\end{enumerate}

\subsection{Framework Overview}
\label{subsec:overview}

The overall architecture of the proposed framework is shown in \cref{fig:pipeline}. To condition the generation process on both anatomy and disease severity, our architecture first combines an IP-Adapter with an AOE. To resolve the entanglement problem defined in \cref{subsec:problem}, we integrate three novel components into this combined baseline: a Feature Purifier, Triple-Pathway Cross-Attention modules, and a delta steering mechanism.

During training, both the VAE encoder/decoder~\cite{rombach2022high} and the CLIP ViT-L/14~\cite{radford2021learning} image encoder are kept frozen. Given an endoscopic image $\mat{I} \in \mathbb{R}^{3 \times 256 \times 256}$, we extract CLIP features and project them to $N{=}16$ tokens via a trainable Perceiver Resampler, yielding $\eimg \in \mathbb{R}^{N \times D}$ with $D{=}768$. These tokens carry both anatomical structure and the disease state.

Simultaneously, the target Mayo Endoscopic Score $y_t$ is encoded by the AOE into disease tokens $\eaoe^{\mathrm{t}} \in \mathbb{R}^{N \times D}$.
The complete conditioning context fed to each U-Net cross-attention layer concatenates three token groups:

\begin{equation}
\mat{C} = \big[\,\underbrace{\eaoe^{\mathrm{t}}}_{N}\;\|\;\underbrace{\eclean}_{N}\;\|\;\underbrace{\edelta}_{N}\,\big]
\in \mathbb{R}^{3N \times D}
\label{eq:conditioning}
\end{equation}

where $\eclean$ denotes the purified anatomy tokens obtained via the proposed feature purifier (\cref{subsec:purifier}) and $\edelta$ encodes the directional severity change.
During training, both source and target labels are identical ($y_s{=}y_t$), so $\edelta{=}\mathbf{0}$ and the model learns standard image reconstruction. At inference, a different target activates the delta pathway to steer disease generation (\cref{subsec:delta}).

\subsection{Additive Ordinal Embedder (AOE)}
\label{subsec:aoe}

AOE mirrors the clinical observation that a higher MES level exhibits all features of lower levels plus additional pathology and represents severity as a cumulative sum of learnable increments~(\cref{eq:aoe}).\looseness=-1
\begin{equation}
E[k] = \vecb{b} + \sum_{i=0}^{k-1} \boldsymbol{\delta}_i, \quad k \in \{0, \ldots, K{-}1\}
\label{eq:aoe}
\end{equation}

where $\vecb{b} \in \mathbb{R}^D$ is a learnable base vector representing healthy mucosa and each $\boldsymbol{\delta}_i$ captures the incremental change associated with increasing severity (initialized with positive mean to encourage monotonic progression).
Continuous interpolation $E(y) = (1{-}\alpha)\,E[\lfloor y \rfloor] + \alpha\,E[\lceil y \rceil]$ enables generation at arbitrary intermediate severity levels.
The resulting embedding is projected to $N$ tokens via a two-layer MLP with GELU activation and LayerNorm.

\subsection{Feature Purifier}
\label{subsec:purifier}

\begin{figure*}[t]
  \centering
  \includegraphics[width=0.97\linewidth]{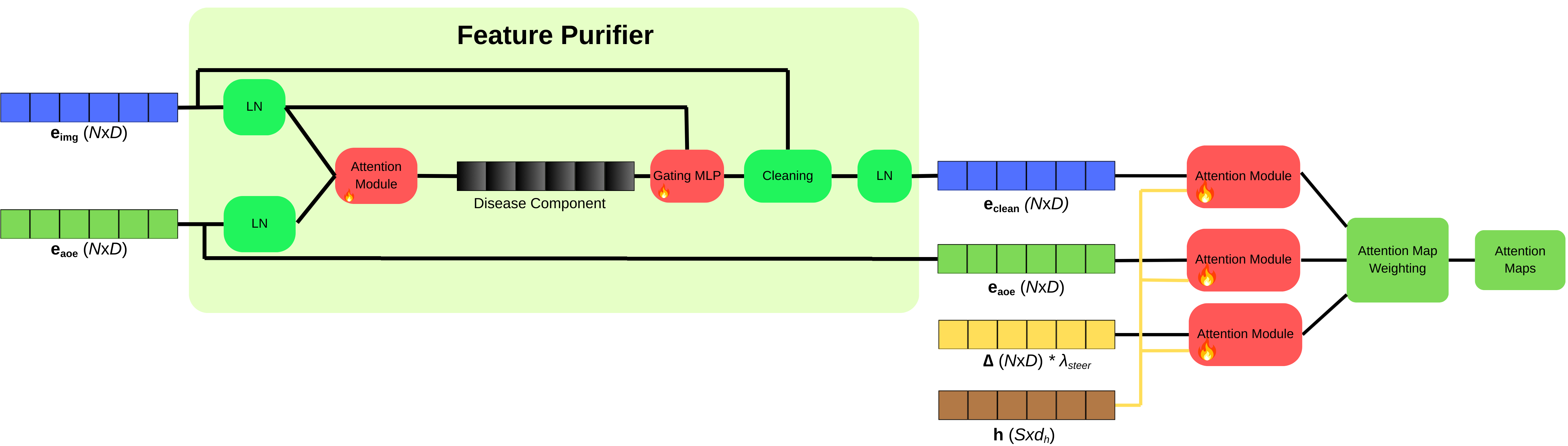}
  \caption{\textbf{Triple Pathway Cross-Attention:} All pathways share query $\mat{Q} = W_Q\,\mat{h}$ from U-Net hidden states $\mat{h}$. The anatomy pathway applies pretrained $W_K, W_V$ to purified tokens $\mat{e}_{\mathrm{clean}}$. The disease pathway applies separate bias-free projections $W_K^{\mathrm{d}}, W_V^{\mathrm{d}}$ (warm-started from $W_K, W_V$) to AOE tokens $\mat{e}_{\mathrm{aoe}}^{\mathrm{t}}$. The delta pathway (inference only) processes the ordinal difference $\Delta$ using these same bias-free projections, ensuring $\Delta{=}\mathbf{0}$ yields zero output during training. Outputs are combined via resolution-aware routing gates. \textbf{Feature Purifier:} Cross-attention identifies disease $\vecb{d}$; a gating MLP generates a sigmoid mask $\vecb{g}$; gated residual subtraction isolates anatomy tokens $\eclean$.}
\label{fig:attention_detail}
\end{figure*}

The CLIP image tokens $\eimg$ encode both the patient's anatomy and the current disease state.
When the target severity differs from the source, this entanglement introduces conflicting signals: the model would simultaneously try to preserve the existing disease appearance and generate a different one. The Feature Purifier addresses this by selectively removing disease-correlated information while retaining anatomical structure~(\cref{fig:attention_detail}) in three stages. First, a cross-attention layer queries the image tokens against the source disease embedding ($\eaoe^{\mathrm{s}}$) to identify the disease component in this particular image (\cref{eq:purifier_crossattn}), where LN is Layer Normalization.

\begin{equation}
\vecb{d} = \mathrm{CrossAttn}\!\bigl(Q{=}\mathrm{LN}(\eimg),\; K{=}V{=}\mathrm{LN}(\eaoe^{\mathrm{s}})\bigr) 
\label{eq:purifier_crossattn}
\end{equation}

Next, a gating MLP takes the concatenation of the detected disease component and the original tokens, producing a per-dimension mask that controls how much of each feature dimension to suppress (\cref{eq:purifier_gate}). Finally, gated residual subtraction yields the cleaned tokens (\cref{eq:purifier_clean}).

\begin{equation}
\vecb{g} = \sigma\!\bigl(\mathrm{MLP}_{2D \to D}([\vecb{d}\,;\,\mathrm{LN}(\eimg)])\bigr) \in (0,1)^{N \times D}
\label{eq:purifier_gate}
\end{equation}

\begin{equation}
\eclean = \mathrm{LN}\!\left(\eimg - \vecb{g} \odot \vecb{d}\right)
\label{eq:purifier_clean}
\end{equation}

Since training uses $y_s{=}y_t$, the purifier learns to identify disease-correlated dimensions from the reconstruction signal alone.
At inference, providing the true source label enables disease removal regardless of the target. Our current formulation assumes that the source MES label $y_s$ is available at inference time.

\subsection{Triple Pathway Cross-Attention}
\label{subsec:routing}
We replace standard U-Net cross-attention with a split-injection processor, using dedicated pathways rather than concatenating all conditioning tokens~(\cref{fig:attention_detail}). 
The U-Net hidden states $\mat{h} \in \mathbb{R}^{S \times d_h}$, where $S$ is the spatial token count and $d_h \in \{320, 640, 1280\}$ is the block's hidden dimension for resolutions $32^2$, $16^2$, $8^2$ respectively, provide the shared query $\mat{Q}=W_Q\,\mat{h}$.
The conditioning tokens, $\eclean,\,\eaoe^{\mathrm{t}},\,\edelta \in \mathbb{R}^{N \times D}$ with $N{=}16$, $D{=}768$, are projected into $d_h$ dimensions by independent key-value projections.
Each pathway attends separately:

\begin{align}
\mathbf{z}_{\mathrm{a}} &= \mathrm{softmax}\left(\frac{\mathbf{Q}\,(W_K\,\mathbf{e}_{\mathrm{clean}})^{\top}}{\sqrt{d_h}}\right)\,W_V\,\mathbf{e}_{\mathrm{clean}} \label{eq:z_anat} \\[3pt]
\mathbf{z}_{\mathrm{d}} &= \mathrm{softmax}\left(\frac{\mathbf{Q}\,(W_K^{\mathrm{d}}\,\mathbf{e}_{\mathrm{aoe}}^{\mathrm{t}})^{\top}}{\sqrt{d_h}}\right)\,W_V^{\mathrm{d}}\,\mathbf{e}_{\mathrm{aoe}}^{\mathrm{t}} \label{eq:z_dis}
\end{align}

The anatomy pathway applies Stable Diffusion's~\cite{rombach2022high} pretrained $W_K, W_V$ (with bias) to $\eclean$, preserving structural reconstruction capacity. The disease pathway applies separate, bias-free projections $W_K^{\mathrm{d}}, W_V^{\mathrm{d}}$ (warm-started from pretrained weights) to $\eaoe$. Crucially, bias removal ensures zero-valued delta tokens yield exactly zero output during training. At inference, the delta pathway processes $\edelta$ through these identical bias-free projections.

\noindent\textbf{Frequency-aware routing gates.}
The relative importance of anatomy versus disease information varies across U-Net resolution levels: low-resolution feature maps capture global structure while high-resolution maps encode fine texture.
We exploit this by assigning each block a resolution-dependent gate pair $(\ganat, \gdis)$ that weights the two pathways.

\noindent\textbf{Output combination.}
The final attention map is obtained after combining all pathways. Fixed routing gates $(\ganat, \gdis)$ weight the anatomy and disease contributions per block; $\lambda_{\text{steer}}$ scales the delta signal as in \cref{eq:combination}.

\begin{equation}
\mathbf{z}^{(\ell)} =
g_{\mathrm a}^{(\ell)} \mathbf{z}_{\mathrm a}^{(\ell)} +
g_{\mathrm d}^{(\ell)} \mathbf{z}_{\mathrm d}^{(\ell)} +
\lambda_{\mathrm{steer}} \mathbf{z}_{\delta}^{(\ell)},
\label{eq:combination}
\end{equation}
where $\lambda_{\text{steer}}{=}0$ during training and the delta pathway activates only at inference to inject $z_{\delta}$ (formally defined in ~\cref{subsec:delta}) and $(g_{\mathrm a}^{(\ell)}, g_{\mathrm d}^{(\ell)})$ denotes the fixed routing pair assigned to U-Net block $\ell$.

\subsection{Delta Steering}
\label{subsec:delta}

CFG is commonly used to strengthen conditioning at inference, but it requires running both conditional and unconditional passes, effectively doubling computational cost. It does not provide an inherent notion of directionality (increasing disease severity vs. decreasing it).
We introduce delta steering, which operates directly at the cross-attention level to provide signed, magnitude-proportional severity control in a single forward pass.

The delta embedding is computed as a post-projection difference, where $\operatorname{proj}(\cdot)$ denotes the two-layer MLP, so that shared biases cancel exactly (\cref{eq:delta}).
\begin{equation}
\edelta = \operatorname{proj}(E[y_t]) - \operatorname{proj}(E[y_s]) \in \mathbb{R}^{N \times D}
\label{eq:delta}
\end{equation}
When $y_t > y_s$ the delta points toward more severe disease; when $y_t < y_s$ it reverses direction.
The magnitude naturally scales with $|y_t{-}y_s|$, providing ordinal proportionality without manual tuning.

At inference, $\edelta$ is processed through the same bias-free disease projections to produce an independent attention contribution:
\begin{equation}
\mathbf{z}_{\delta} = \mathrm{softmax}\left(\frac{\mathbf{Q}\,(W_K^{\mathrm{d}}\,\boldsymbol{\Delta})^{\top}}{\sqrt{d_h}}\right)\,W_V^{\mathrm{d}}\,\boldsymbol{\Delta}
\label{eq:z_delta}
\end{equation}
Three design choices guarantee that this pathway is exactly inactive during training: (i)~post-projection subtraction makes $\edelta{=}\mathbf{0}$ when $y_s{=}y_t$; (ii)~bias-free projections ensure $W_K^{\mathrm{d}}\,\mathbf{0} = \mathbf{0}$; and (iii)~$\lambda_{\text{steer}}{=}0$ provides an explicit gate.

\subsection{Training Objective}
\label{subsec:training_objective}

We train with the standard $\boldsymbol{\epsilon}$-prediction objective~\cite{ho2020denoising}, weighted by Min-SNR-$\gamma$~\cite{hang2023efficient} ($\gamma{=}1$) to balance loss contributions across timesteps.

In the routing-gate variants, neither the image tokens nor the AOE tokens are dropped or zeroed: delta steering provides inference-time control without requiring an unconditional pass.
In ablation configurations in which delta steering and routing gates are removed (IP-Text and IP-AOE), we fall back on CFG, which requires random image-token dropout ($p{=}0.1$) during training.
The AOE negative embedding for CFG is not a null vector but a contrastive severity: healthy images (MES~0) are guided away from mild disease features, while diseased images (MES~${\geq}1$) are guided away from healthy mucosa, with smooth interpolation in between.

\subsection{Downstream Data Augmentation}
\label{subsec:augmentation}

For each training image with source label $y_s$, we generate synthetic images for all other MES levels, effectively transforming the naturally imbalanced training set into a balanced corpus. To measure the downstream clinical utility of this augmentation, a ResNet-18~\cite{he2016deep} classifier is trained on each augmented configuration and evaluated on a held-out test set using accuracy and Quadratic Weighted Kappa (QWK). To explicitly assess target severity fidelity, an independent ResNet-18 regression judge, trained exclusively on real LIMUC data, evaluates the synthetic outputs in isolation. This ensures the generative process genuinely transforms pathology rather than merely replicating the source.

% !TEX root = ../main.tex
\section{Experimental Setup}
\label{sec:experiments}

\noindent \textbf{Dataset.} We use the LIMUC dataset, comprising 11,276 MES-labeled endoscopic images from 564 patients.
Following~\cite{ccauglar2023ulcerative}, 992 images with artifacts are removed; the remaining 10,284 images are cropped to $224{\times}224$ and resized to $256{\times}256$.
The patient-level split yields 7,180 train, 1,661 val, and 1,443 test images, with severe imbalance: MES 0-3 are 54.4\%, 27.0\%, 11.0\%, and 7.6\%, respectively.

\noindent \textbf{Training Configuration.} We fine-tune Stable Diffusion v1.4~\cite{rombach2022high} with frozen VAE and CLIP ViT-L/14; the U-Net, Perceiver Resampler, AOE, and Feature Purifier are trained jointly.
We use AdamW~\cite{loshchilov2018decoupled} ($\text{lr}{=}10^{-4}$, $2{\times}$ multiplier for Resampler and Purifier) with linear warmup (2 epochs) followed by cosine annealing to $10^{-6}$.
Training runs for 150 epochs with a batch size of 64 on NVIDIA RTX 6000 Ada GPUs using FP16, gradient checkpointing, and Exponential Moving Average (EMA) ($\tau{=}0.999$).

The diffusion process uses 1,000 timesteps with linear schedule ($\beta_1{=}8.5{\times}10^{-4}$, $\beta_T{=}0.012$) and Min-SNR-$\gamma$~\cite{hang2023efficient} ($\gamma{=}1.0$). Denoising Diffusion Implicit Models
(DDIM)~\cite{song2022denoisingdiffusionimplicitmodels} with 50 steps is used for inference.
Class-balanced sampling mitigates dataset imbalance.
The AOE uses $K{=}4$ classes, $D{=}768$, $N{=}16$ tokens, and delta init scale~$0.05$.
Augmentation: random flips, rotation ($\pm 5^\circ$), center crop $224{\times}224$, random perspective ($p{=}0.2$).

\subsection{Experimental Design}
\label{subsec:ablation}

We evaluate against two baselines:

\begin{itemize}[nosep, leftmargin=*]
\item \textbf{IP-Adapter with CLIP text encoder (IP-Text):}  IP-Adapter is used with the CLIP text encoder to generate images using severity descriptions as text prompts (e.g., \textit{``MES~$k$''}, $k \in \{0,1,2,3\}$). For a fair comparison, the U-Net is also fine-tuned for this baseline; we observed that utilizing a frozen U-Net was insufficient to capture the specific anatomical and pathological nuances of the endoscopic domain.

\item \textbf{IP-Adapter with AOE (IP-AOE)}: This baseline combines IP-Adapter with the AOE for disease conditioning. IP-Adapter injects CLIP ViT-L/14 image features into the U-Net via decoupled cross-attention, providing anatomical conditioning from a reference endoscopic image. The AOE encodes the target MES label as a cumulative ordinal embedding, which is injected alongside the image features and steered at inference via CFG.
\end{itemize}

To isolate the contribution of routing gate assignment, we compare three configurations that share identical training hyperparameters. All variants include the Feature Purifier and delta steering; they differ only in the gate weights $(g_{\mathrm{a}}, g_{\mathrm{d}})$ per U-Net resolution level.

\begin{itemize}[nosep, leftmargin=*]
\item \textbf{Hierarchical Routing (DADD-H)}: High-resolution blocks ($32^2$, $16^2$) emphasize disease $(g_{\mathrm{d}}{=}0.9,\, g_{\mathrm{a}}{=}0.1)$; low-resolution blocks ($8^2$) emphasize anatomy $(g_{\mathrm{a}}{=}0.9,\, g_{\mathrm{d}}{=}0.1)$. This reflects the U-Net resolution hierarchy: fine-grained layers encode inflammatory texture while coarse layers capture global mucosal structure.

\item \textbf{Inverted Routing (DADD-I)}: Gate assignment reversed relative to Hierarchical Routing, where disease emphasis is on low-resolution blocks, and anatomy emphasis on high-resolution blocks.

\item \textbf{Uniform Routing (DADD-U)}: All blocks use $g_{\mathrm{a}}{=}g_{\mathrm{d}}{=}0.5$.

\end{itemize}

\cref{fig:scale_selection} shows generated images at steering scales $\lambda_{\text{steer}} \in \{1,3,5,7\}$. 
At low scales, disease modulation is subtle and severity levels remain difficult to distinguish; at high scales, artifacts dominate, and anatomical consistency degrades. We therefore select $\lambda_{\text{steer}}{=}3$ as the operating point for all subsequent evaluations.

\subsection{Evaluation Protocol}
\label{subsec:evaluation} 
\noindent\textbf{Generation quality.}
Generation quality is assessed via Fréchet Inception Distance (FID)~\cite{heusel2017fid}, CLIP Maximum Mean Discrepancy (CMMD)~\cite{jayasumana2024cmmd}, and improved Precision and Recall~\cite{Kynknniemi2019ImprovedPA}, which measure how closely the synthetic distribution matches real data. Images are generated across all four MES classes using both LIMUC validation and testing images as anatomy conditions. This combined conditioning set is utilized to ensure a sufficiently large sample size for robust statistical estimation of the distribution-based metrics. The IP-Text and IP-AOE are steered via CFG guidance scale $w {=} 3$, while the other variants use attention-level delta steering scale $\lambda_{\text{steer}} {=} 3$.

\noindent\textbf{Downstream classification.} For each variant, we produce a balanced augmented training set (\cref{subsec:augmentation}) by combining real training images with synthetic samples. A ResNet-18 classifier is trained on this augmented set for 30 epochs and evaluated on the held-out test set to report accuracy and QWK.

\noindent\textbf{Regression Judge.} To directly assess whether synthetic images reflect their target severity, we employ a Regression Judge: a ResNet-18 trained exclusively on real LIMUC training data to predict MES as a continuous value. The judge is evaluated solely on synthetic images generated using the held-out test set as anatomical conditions, with target severity interpolated from MES 0 to 3 in increments of 0.2. Performance is reported using QWK and Root Mean Square Error (RMSE) against these target labels, with QWK computed after clipping predictions to $[0,3]$ and discretizing both predictions and targets to the nearest MES class. A model that genuinely transforms the appearance of disease will score highly; a model producing near-copies of the source will score near zero, regardless of FID.

\noindent\textbf{Feature-Space Analysis.}
To verify overlap between real and synthetic distributions, we compute the global Silhouette Score~\cite{ROUSSEEUW198753} (using cosine distance) directly on the 512-d ResNet-18 penultimate features. UMAP~\cite{McInnes2018UMAPUM} is utilized strictly for 2-D qualitative visualization.

\begin{figure*}[t]
  \centering
  \includegraphics[width=0.70\linewidth]{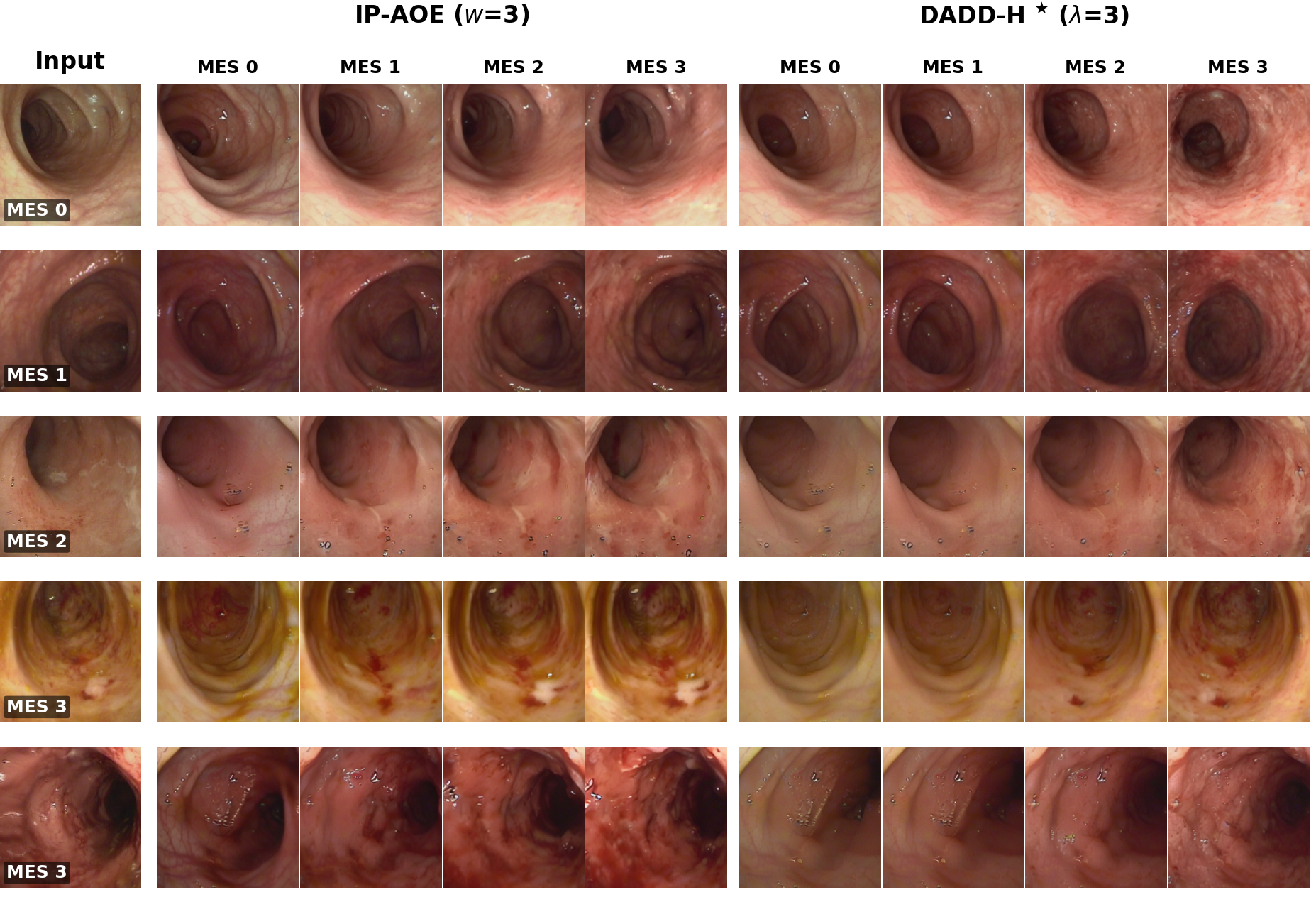}
  \caption{\textbf{Qualitative comparison: IP-AOE vs.\ DADD-H.}
  Each row shows a source image (MES~0-3, input at left) generated at four target MES levels by the IP-AOE and DADD-H.}
  \label{fig:model_comparison}
\end{figure*}

% !TEX root = ../main.tex
\section{Results}
\label{sec:results}

\begin{figure}[t]
  \centering
  \includegraphics[width=0.8\linewidth]{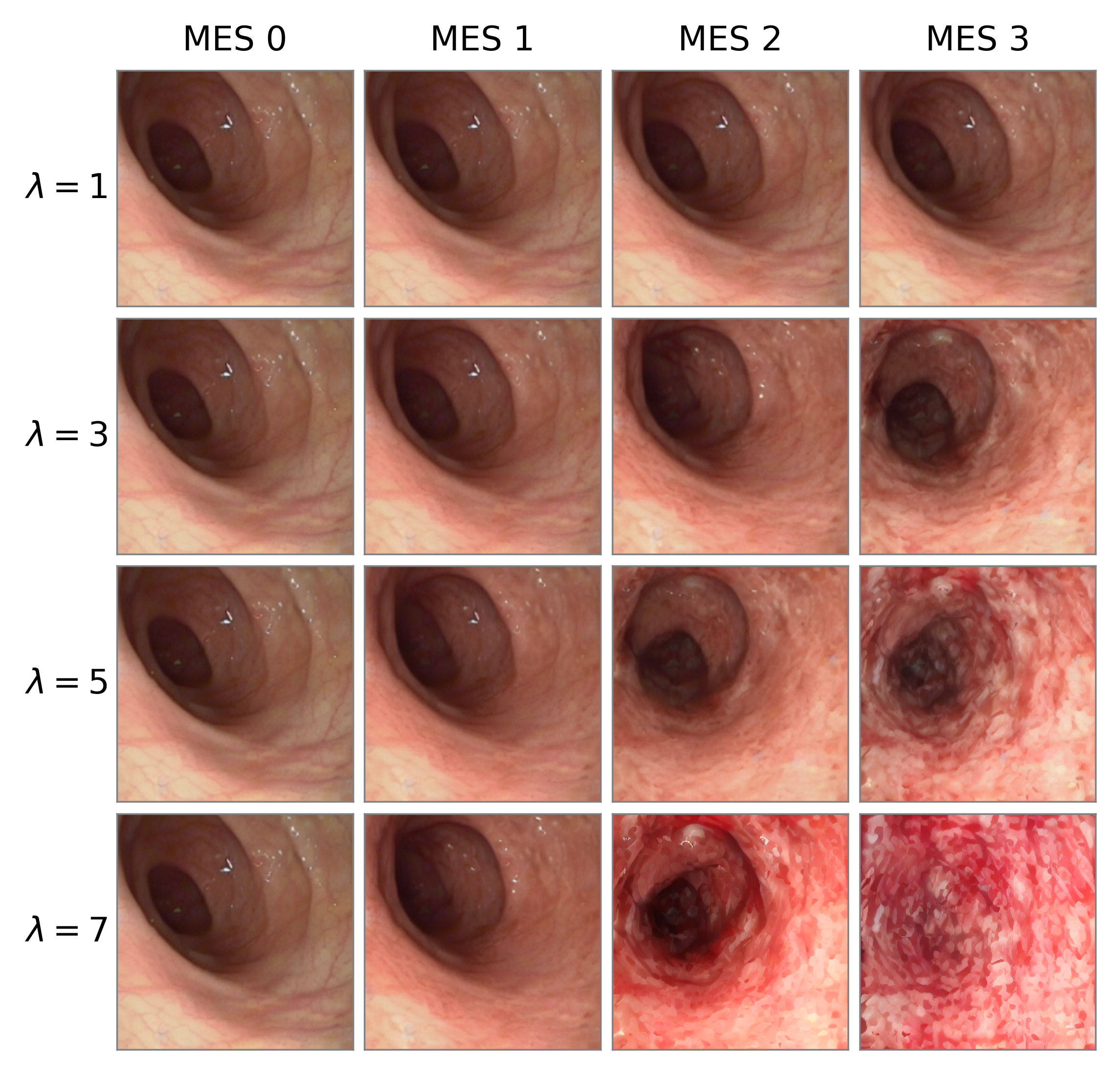}
  \caption{\textbf{Steering scale selection.}
  A MES~0 source image generated by DADD-H at various steering scales across target MES levels. At $\lambda_{\text{steer}}{=}3$, the model produces clearly distinguishable severity levels while preserving anatomical consistency. Lower scales under-modulate and higher scales introduce artifacts.}
   \label{fig:scale_selection}
\end{figure}

\noindent\textbf{Quantitative Evaluation.}
As shown in \Cref{tab:gen_quality}, the DADD-I and DADD-U configurations achieve lower (better) FID and CMMD scores than DADD-H. However, these metrics primarily measure distributional similarity rather than the faithfulness of disease transformation.  A model that produces near-copies of the conditioning image achieves low FID by construction, as its outputs inherently match the real data distribution regardless of disease fidelity. Consequently, lower FID scores for DADD-I indicate under-modulation rather than higher quality synthesis. This outcome validates our architectural premise: fine-grained layers are responsible for encoding localized inflammatory textures, whereas coarse layers capture global mucosal structure. By reversing this injection strategy, DADD-I fails to adequately alter pathological features, resulting in severely under-modulated disease representations. Crucially, as established in our scale selection (\cref{fig:scale_selection}), attempting to compensate for this under-modulation by increasing the steering scale ($\lambda_{steer} > 3$) does not recover the missing pathological textures, but rather induces severe anatomical collapse, confirming the necessity of correct frequency routing. We therefore jointly interpret FID and CMMD with the downstream classification results (see \cref{tab:merged_gen_quality}), which directly assess whether the synthesized disease features are discriminative.

\begin{table}[t]
\centering
\small
\setlength{\tabcolsep}{4pt}
\caption{\textbf{Generation quality of baseline methods and DADD across configurations.} Improved Precision and Recall~\cite{Kynknniemi2019ImprovedPA} metrics were calculated using a neighborhood size of $k=3$ on random subsets of up to 10,000 samples per distribution.}
\label{tab:gen_quality}
\begin{tabular}{@{}l cccc@{}}
\toprule
\textbf{Variant} & \textbf{FID}$\downarrow$ & \textbf{CMMD}$\downarrow$ & \textbf{Prec}$\uparrow$ & \textbf{Rec}$\uparrow$ \\
\midrule
IP-Text  & {35.60} & {.033} & {.736} & {.417}  \\
IP-AOE          & 25.36 & .033 & .606 & \textbf{.584} \\
DADD-H & 27.20 & .037 & .692 & .463 \\
DADD-I          & \textbf{23.46} & .032 & .760 & .518 \\
DADD-U            & 23.69 & \textbf{.032} & \textbf{.768} & .521 \\
\bottomrule
\end{tabular}

\end{table}

\begin{table}[t]
\centering
\scriptsize
\setlength{\tabcolsep}{5pt}
\caption{\textbf{Downstream Evaluation.} Classification performance (mixed vs. synthetic-only data) and regression-based severity fidelity. For the synthetic-only task, training sets are balanced by generating all MES levels per image and downsampled to the minority class size. Values for synthetic-only data denote mean $\pm$ standard deviation across 5 runs.}
\label{tab:merged_gen_quality}
\begin{tabular}{@{}l cc cc cc@{}}
\toprule
& \multicolumn{2}{c}{\textbf{Mixed Training}} & \multicolumn{2}{c}{\textbf{Synthetic Only}} & \multicolumn{2}{c}{\textbf{Regression Judge}} \\
\cmidrule(lr){2-3} \cmidrule(lr){4-5} \cmidrule(lr){6-7}
\textbf{Variant} & \textbf{Acc}$\uparrow$ & \textbf{QWK}$\uparrow$ & \textbf{Acc}$\uparrow$ & \textbf{QWK}$\uparrow$ & \textbf{QWK}$\uparrow$ & \textbf{RMSE}$\downarrow$ \\
\midrule
\rowcolor{gray!10}
\textit{Real data} & 74.84 & 83.14 & --- & --- & 83.28 & 0.47 \\
\midrule
IP-Text & 76.10 & 84.46 & 36.42 ± 2.50 & 13.89 ± 8.90 & 15.44 & 1.26 \\
IP-AOE  & 76.44 & \textbf{84.53} & 54.54 ± 1.23 & 42.20 ± 1.51 & 54.53 & 0.80 \\
DADD-H  & 76.30 & 84.50 & \textbf{61.68 ± 2.38} & \textbf{70.74 ± 1.80} & \textbf{81.53} & \textbf{0.46} \\
DADD-I  & \textbf{76.51} & 83.34 & 25.69 ± 0.44 & 24.55 ± 5.53 & 22.14 & 1.10 \\
DADD-U  & 73.66 & 83.00 & 19.25 ± 0.70 & 16.77 ± 5.87 & 17.09 & 1.19 \\
\bottomrule
\end{tabular}
\end{table}

\cref{tab:merged_gen_quality} details downstream classification (via class-balanced \textit{Mixed Training} or \textit{Synthetic Only} data) and severity fidelity (\textit{Regression Judge}). The shaded reference row denotes baselines trained strictly on imbalanced real data using cross-entropy (CE) loss, which outperformed Focal Loss (74.15 Acc, 82.08 QWK).
Most variants improve classification accuracy over the imbalanced baseline, confirming the benefit of synthetic balancing. However, the `Synthetic Only' results provide a more rigorous test. While DADD-H enables a classifier to reach 61.68\% accuracy without seeing real data, outperforming the IP-AOE baseline (54.54\%), the near-random performance of DADD-I and DADD-U ($<$26.0\%) confirms they fail to synthesize discriminative pathological features. Similarly, the failure of the IP-Text baseline (13.89 QWK) underscores that generic text embeddings lack the granularity required for monotonic progression. 
The Regression Judge provides the final validation of severity fidelity. DADD-H achieves 81.53 QWK / 0.46 RMSE, approaching the real-data reference (83.28 / 0.47) and confirming that synthesized images strongly align with intended MES levels. This dissociation between FID and severity fidelity confirms that distributional similarity alone is insufficient to validate conditional progression synthesis. By contrast, the IP-AOE baseline and inverted/uniform variants achieve significantly lower fidelity (54.53, 22.14, and 17.09 QWK, respectively) despite achieving lower (better) FID than DADD-H.
\begin{table}[t]
\centering
\scriptsize
\setlength{\tabcolsep}{2pt}
\caption{\textbf{Inference speed under a fair protocol} on NVIDIA RTX 6000 Ada (DDIM-50, 256$\times$256). All models are evaluated with the same precision (FP16), and timing includes \emph{generation only} (model loading excluded).}
\label{tab:speed_benchmark}
\begin{tabular}{@{}lcccc@{}}
\toprule
\textbf{Variant} & \textbf{ms/img}$\downarrow$ (BS=1) & \textbf{ms/img}$\downarrow$ (BS=8) & \textbf{Img/s}$\uparrow$ (BS=8) & \textbf{Mem (GB)}$\downarrow$ (BS=8) \\
\midrule
IP-Text & 2766.5 & 351.2 & 2.85 & \textbf{7.16} \\
IP-AOE  & 2442.0 & 314.6 & 3.18 & 8.02 \\
DADD-H  & \textbf{1432.6} & \textbf{187.3} & \textbf{5.34} & 8.17 \\
DADD-I  & 1437.9 & 189.5 & 5.28 & 8.17 \\
DADD-U  & 1476.5 & 196.2 & 5.10 & 8.17 \\
\bottomrule
\end{tabular}
\end{table}

\noindent\textbf{Inference Speed.} \cref{tab:speed_benchmark} demonstrates that all DADD variants significantly outperform the baselines in generation speed. By utilizing single-pass delta steering instead of the two-branch CFG computation required by IP-Text and IP-AOE, DADD achieves roughly a $2\times$ speedup (187.3 ms/img for DADD-H at batch size (BS)=8), trading a marginal increase in memory footprint for substantially faster inference.

\begin{figure}[t]
  \centering
  \includegraphics[width=0.9\linewidth]{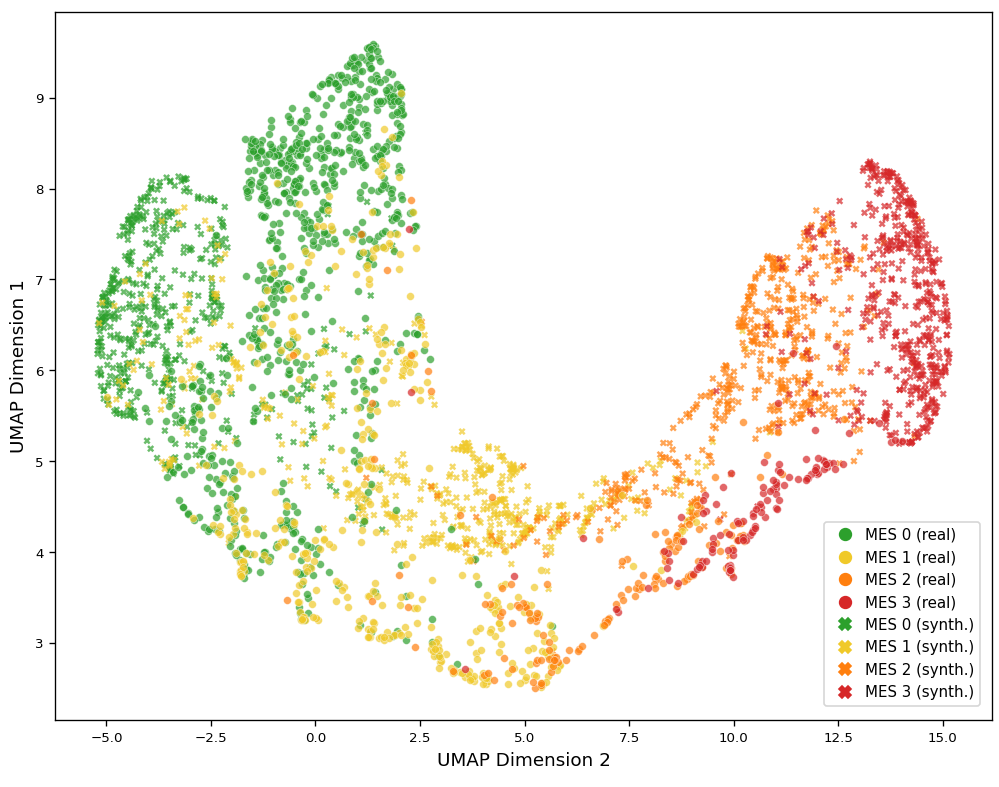}
  \caption{\textbf{Feature-space alignment.}
  UMAP of real~($\bullet$) vs.\ synthetic~($\times$) images (DADD-H, $\lambda_{\text{steer}}{=}3$). The global Silhouette Score of 0.1296 confirms overlap.}
  \label{fig:umap_global}
\end{figure}

\noindent\textbf{Feature-Space Analysis.}
To quantify distribution separability, we compute the global Silhouette Score (cosine distance) on the native 512-d ResNet-18 features. \cref{fig:umap_global} confirms real-synthetic overlap, supported by a low score of~0.1296, alongside a 2-D UMAP~\cite{McInnes2018UMAPUM} visualization.

\noindent\textbf{Qualitative Model Comparison.}
As illustrated in \cref{fig:model_comparison}, our model yields clearly distinguishable transitions in severity while largely preserving instance-specific anatomy. Crucially, when synthesizing lower severity levels from a diseased source (e.g., MES~3 $\to$ 0), our approach successfully erases existing pathological textures. In contrast, IP-AOE fails to remove the source disease embedded in the IP-Adapter tokens, resulting in entangled generations in which healthy mucosa is incorrectly superimposed on existing inflamed structures.
% !TEX root = ../main.tex
\section{Conclusion}
We presented DADD, a disentangled conditioning framework for synthesizing endoscopic disease progression. By utilizing a Feature Purifier to erase source disease from CLIP tokens and frequency-aware routing gates to direct anatomical and pathological signals to appropriate U-Net layers, our model enables a plausible visualization of that specific individual's bidirectional disease trajectory, accurately simulating both progressive inflammation and mucosal healing. We mitigated the semantic conflicts inherent in standard adapters and suppressed existing pathological textures from the condition image. Our delta steering mechanism further enables precise, single-pass severity control. Validated on the LIMUC dataset, our hierarchical model configuration (DADD-H) achieved a QWK of 81.53, closely matching the real-data ceiling of 83.28 on the regression judge task. We demonstrated that synthetic balancing consistently improves classification accuracy over imbalanced baselines, and that low FID scores can be misleading as they fail to assess faithful pathological transformation. Despite these gains, model performance remains sensitive to the steering scale $\lambda_{steer}$, where values $\lambda_{steer} \ge 5$ can introduce structural artifacts. Future research will explore per-class steering schedules to optimize transitions across severity levels and extend the DADD architecture to other longitudinal medical modalities.
{
    \small
    \bibliographystyle{ieeenat_fullname}
    \bibliography{main}

@String(CVPR= {IEEE Conf. Comput. Vis. Pattern Recog.})

@String(NIPS= {Adv. Neural Inform. Process. Syst.})

@String(ICLR = {Int. Conf. Learn. Represent.})

@String(AAAI = {AAAI})

@String(CVPR  = {CVPR})

@String(NIPS  = {NeurIPS})

@String(ICLR  = {ICLR})

@inproceedings{ho2020denoising,
author = {Ho, Jonathan and Jain, Ajay and Abbeel, Pieter},
title = {Denoising diffusion probabilistic models},
year = {2020},
isbn = {9781713829546},
booktitle = {34th International Conference on Neural Information Processing Systems},
articleno = {574},
}

@inproceedings{rombach2022high,
  author={Rombach, Robin and Blattmann, Andreas and Lorenz, Dominik and Esser, Patrick and Ommer, Björn},
  booktitle={2022 IEEE/CVF Conference on Computer Vision and Pattern Recognition (CVPR)}, 
  title={High-Resolution Image Synthesis with Latent Diffusion Models}, 
  year={2022},
  volume={},
  number={},
  pages={10674-10685},
  doi={10.1109/CVPR52688.2022.01042}}

@article{dhariwal2021diffusion,
  title={Diffusion models beat gans on image synthesis},
  author={Dhariwal, Prafulla and Nichol, Alexander},
  journal={Advances in neural information processing systems},
  volume={34},
  pages={8780--8794},
  year={2021}
}

@article{ho2022classifier,
  title={Classifier-free diffusion guidance},
  author={Ho, Jonathan and Salimans, Tim},
  journal={arXiv preprint arXiv:2207.12598},
  year={2022}
}

@inproceedings{song2022denoisingdiffusionimplicitmodels,
title={Denoising Diffusion Implicit Models},
author={Jiaming Song and Chenlin Meng and Stefano Ermon},
booktitle={International Conference on Learning Representations},
year={2021},
url={https://openreview.net/forum?id=St1giarCHLP}
}

@inproceedings{radford2021learning,
  title={Learning Transferable Visual Models From Natural Language Supervision},
  author={Alec Radford and Jong Wook Kim and Chris Hallacy and Aditya Ramesh and Gabriel Goh and Sandhini Agarwal and Girish Sastry and Amanda Askell and Pamela Mishkin and Jack Clark and Gretchen Krueger and Ilya Sutskever},
  booktitle={International Conference on Machine Learning},
  year={2021},
  url={https://api.semanticscholar.org/CorpusID:231591445}
}

@article{ye2023ip,
  title={Ip-adapter: Text compatible image prompt adapter for text-to-image diffusion models},
  author={Ye, Hu and Zhang, Jun and Liu, Sibo and Han, Xiao and Yang, Wei},
  journal={arXiv preprint arXiv:2308.06721},
  year={2023}
}

@inproceedings{jaegle2021perceiver,
  title={Perceiver: General perception with iterative attention},
  author={Jaegle, Andrew and Gimeno, Felix and Brock, Andy and Vinyals, Oriol and Zisserman, Andrew and Carreira, Joao},
  booktitle={International conference on machine learning},
  pages={4651--4664},
  year={2021},
  organization={PMLR}
}

@article{polat2022labeled,
    author = {Polat, Gorkem and Kani, Haluk Tarik and Ergenc, Ilkay and Ozen Alahdab, Yesim and Temizel, Alptekin and Atug, Ozlen},
    title = {Improving the Computer-Aided Estimation of Ulcerative Colitis Severity According to Mayo Endoscopic Score by Using Regression-Based Deep Learning},
    journal = {Inflammatory Bowel Diseases},
    volume = {29},
    number = {9},
    pages = {1431-1439},
    year = {2022},
    month = {11},
    issn = {1078-0998},
    doi = {10.1093/ibd/izac226},
    url = {https://doi.org/10.1093/ibd/izac226},
}

@article{kurt2025progressive,
AUTHOR = {Kurt, Meryem Mine and Çağlar, Ümit Mert and Temizel, Alptekin},
TITLE = {Progressive Disease Image Generation with Ordinal-Aware Diffusion Models},
JOURNAL = {Diagnostics},
VOLUME = {15},
YEAR = {2025},
NUMBER = {20},
ARTICLE-NUMBER = {2558},
URL = {https://www.mdpi.com/2075-4418/15/20/2558},
PubMedID = {41153231},
ISSN = {2075-4418},
DOI = {10.3390/diagnostics15202558}
}

@inproceedings{ccauglar2023ulcerative,
author = { Caglar, Umit Mert and Inci, Alperen and Hanoglu, Oguz and Polat, Gorkem and Temizel, Alptekin },
booktitle = { 2023 IEEE International Conference on Bioinformatics and Biomedicine (BIBM) },
title = {{ Ulcerative Colitis Mayo Endoscopic Scoring Classification with Active Learning and Generative Data Augmentation }},
year = {2023},
volume = {},
ISSN = {},
pages = {462-467},
doi = {10.1109/BIBM58861.2023.10385621},
url = {https://doi.ieeecomputersociety.org/10.1109/BIBM58861.2023.10385621},
publisher = {IEEE Computer Society},
address = {Los Alamitos, CA, USA},
month =Dec}

@inproceedings{he2016deep,
  author={He, Kaiming and Zhang, Xiangyu and Ren, Shaoqing and Sun, Jian},
  booktitle={2016 IEEE Conference on Computer Vision and Pattern Recognition (CVPR)}, 
  title={Deep Residual Learning for Image Recognition}, 
  year={2016},
  volume={},
  number={},
  pages={770-778},
  doi={10.1109/CVPR.2016.90}}

@inproceedings{loshchilov2018decoupled,
title={Decoupled Weight Decay Regularization},
author={Ilya Loshchilov and Frank Hutter},
booktitle={International Conference on Learning Representations},
year={2019},
url={https://openreview.net/forum?id=Bkg6RiCqY7},
}

@inproceedings{heusel2017fid,
  title={GANs Trained by a Two Time-Scale Update Rule Converge to a Local Nash Equilibrium},
  author={Martin Heusel and Hubert Ramsauer and Thomas Unterthiner and Bernhard Nessler and Sepp Hochreiter},
  booktitle={Neural Information Processing Systems},
  year={2017},
  url={https://api.semanticscholar.org/CorpusID:326772}
}

@inproceedings{jayasumana2024cmmd,
  author={Jayasumana, Sadeep and Ramalingam, Srikumar and Veit, Andreas and Glasner, Daniel and Chakrabarti, Ayan and Kumar, Sanjiv},
  booktitle={2024 IEEE/CVF Conference on Computer Vision and Pattern Recognition (CVPR)}, 
  title={Rethinking FID: Towards a Better Evaluation Metric for Image Generation}, 
  year={2024},
  volume={},
  number={},
  pages={9307-9315},
  doi={10.1109/CVPR52733.2024.00889}}

@inproceedings{hang2023efficient,
  title={Efficient diffusion training via min-snr weighting strategy},
  author={Hang, Tiankai and Gu, Shuyang and Li, Chen and Bao, Jianmin and Chen, Dong and Hu, Han and Geng, Xin and Guo, Baining},
  booktitle={Proceedings of the IEEE/CVF international conference on computer vision},
  pages={7441--7451},
  year={2023},
  doi={10.1109/ICCV51070.2023.00684}
}

@inproceedings{qi2024deadiff,
  author={Qi, Tianhao and Fang, Shancheng and Wu, Yanze and Xie, Hongtao and Liu, Jiawei and Chen, Lang and He, Qian and Zhang, Yongdong},
  booktitle={2024 IEEE/CVF Conference on Computer Vision and Pattern Recognition (CVPR)}, 
  title={DEADiff: An Efficient Stylization Diffusion Model with Disentangled Representations}, 
  year={2024},
  volume={},
  number={},
  pages={8693-8702},
  doi={10.1109/CVPR52733.2024.00830}}

@article{xing2025inv,
  title={Inv-Adapter: ID Customization Generation via Image Inversion and Lightweight Parameter Adapter},
  author={Peng Xing and Ning Wang and Jianbo Ouyang and Zechao Li},
  journal={IEEE Transactions on Pattern Analysis and Machine Intelligence},
  year={2025},
  volume={47},
  pages={9938-9952},
  url={https://api.semanticscholar.org/CorpusID:280288114}
}

@article{belrose2023leace,
  title={LEACE: Perfect linear concept erasure in closed form},
  author={Nora Belrose and David Schneider-Joseph and Shauli Ravfogel and Ryan Cotterell and Edward Raff and Stella Biderman},
  journal={ArXiv},
  year={2023},
  volume={abs/2306.03819},
  url={https://api.semanticscholar.org/CorpusID:259088549}
}

@article{han2025groce,
  title={GrOCE: Graph-Guided Online Concept Erasure for Text-to-Image Diffusion Models},
  author={Han, Ning and Ge, Zhenyu and Han, Feng and Sun, Yuhua and Li, Chengqing and Chen, Jingjing},
  journal={arXiv preprint arXiv:2511.12968},
  year={2025}
}

@inproceedings{nguyen2025suma,
  title={SuMa: A Subspace Mapping Approach for Robust and Effective Concept Erasure in Text-to-Image Diffusion Models},
  author={Nguyen, Kien and Tran, Anh and Pham, Cuong},
  booktitle={Proceedings of the IEEE/CVF International Conference on Computer Vision},
  pages={19587--19596},
  year={2025}
}

@inproceedings{hertz2022prompt,
  author       = {Amir Hertz and
                  Ron Mokady and
                  Jay Tenenbaum and
                  Kfir Aberman and
                  Yael Pritch and
                  Daniel Cohen{-}Or},
  title        = {Prompt-to-Prompt Image Editing with Cross-Attention Control},
  booktitle    = {The Eleventh International Conference on Learning Representations,
                  {ICLR} 2023, Kigali, Rwanda, May 1-5, 2023},
  publisher    = {OpenReview.net},
  year         = {2023},
  url          = {https://openreview.net/forum?id=\_CDixzkzeyb},
}

@article{gaintseva2025casteer,
  title={CASteer: Cross-Attention Steering for Controllable Concept Erasure},
  author={Gaintseva, Tatiana and Oncescu, Andreea-Maria and Ma, Chengcheng and Liu, Ziquan and Benning, Martin and Slabaugh, Gregory and Deng, Jiankang and Elezi, Ismail},
  journal={arXiv preprint arXiv:2503.09630},
  year={2025}
}

@inproceedings{nguyen2025supercharged,
  title={Supercharged One-step Text-to-Image Diffusion Models with Negative Prompts},
  author={Nguyen, Viet and Nguyen, Anh and Dao, Trung and Nguyen, Khoi and Pham, Cuong and Tran, Toan and Tran, Anh},
  booktitle={Proceedings of the IEEE/CVF International Conference on Computer Vision},
  pages={18004--18013},
  year={2025}
}

@inproceedings{zhang2023adding,
  title={Adding conditional control to text-to-image diffusion models},
  author={Zhang, Lvmin and Rao, Anyi and Agrawala, Maneesh},
  booktitle={Proceedings of the IEEE/CVF international conference on computer vision},
  pages={3836--3847},
  year={2023}
}

@article{chambon2022roentgen,
  title={Roentgen: vision-language foundation model for chest x-ray generation},
  author={Chambon, Pierre and Bluethgen, Christian and Delbrouck, Jean-Benoit and Van der Sluijs, Rogier and Po{\l}acin, Ma{\l}gorzata and Chaves, Juan Manuel Zambrano and Abraham, Tanishq Mathew and Purohit, Shivanshu and Langlotz, Curtis P and Chaudhari, Akshay},
  journal={arXiv preprint arXiv:2211.12737},
  year={2022}
}

@inproceedings{aversa2024diffinfinite,
author = {Aversa, Marco and Nobis, Gabriel and H\"{a}gele, Miriam and Standvoss, Kai and Chirica, Mihaela and Murray-Smith, Roderick and Alaa, Ahmed and Ruff, Lukas and Ivanova, Daniela and Samek, Wojciech and Klauschen, Frederick and Sanguinetti, Bruno and Oala, Luis},
title = {DiffInfinite: large mask-image synthesis via parallel random patch diffusion in histopathology},
year = {2023},
publisher = {Curran Associates Inc.},
address = {Red Hook, NY, USA},
booktitle = {Proceedings of the 37th International Conference on Neural Information Processing Systems},
articleno = {3413},
numpages = {16},
location = {New Orleans, LA, USA},
series = {NIPS '23}
}

@inproceedings{shin2018medical,
author="Shin, Hoo-Chang
and Tenenholtz, Neil A.
and Rogers, Jameson K.
and Schwarz, Christopher G.
and Senjem, Matthew L.
and Gunter, Jeffrey L.
and Andriole, Katherine P.
and Michalski, Mark",
editor="Gooya, Ali
and Goksel, Orcun
and Oguz, Ipek
and Burgos, Ninon",
title="Medical Image Synthesis for Data Augmentation and Anonymization Using Generative Adversarial Networks",
booktitle="Simulation and Synthesis in Medical Imaging",
year="2018",
isbn="978-3-030-00536-8"
}

@article{trabucco2023effective,
  title={Effective Data Augmentation With Diffusion Models},
  author={Brandon Trabucco and Kyle Doherty and Max Gurinas and Ruslan Salakhutdinov},
  journal={ArXiv},
  year={2023},
  volume={abs/2302.07944},
  url={https://api.semanticscholar.org/CorpusID:256900870}
}

@article{McInnes2018UMAPUM,
  title={UMAP: Uniform Manifold Approximation and Projection for Dimension Reduction},
  author={Leland McInnes and John Healy},
  journal={ArXiv},
  year={2018},
  volume={abs/1802.03426},
  url={https://api.semanticscholar.org/CorpusID:3641284}
}

@article{frid2018gan,
title = {GAN-based synthetic medical image augmentation for increased CNN performance in liver lesion classification},
journal = {Neurocomputing},
volume = {321},
pages = {321-331},
year = {2018},
issn = {0925-2312},
doi = {https://doi.org/10.1016/j.neucom.2018.09.013},
url = {https://www.sciencedirect.com/science/article/pii/S0925231218310749},
author = {Maayan Frid-Adar and Idit Diamant and Eyal Klang and Michal Amitai and Jacob Goldberger and Hayit Greenspan},
}

@article{yi2019generative,
title = {Generative adversarial network in medical imaging: A review},
journal = {Medical Image Analysis},
volume = {58},
pages = {101552},
year = {2019},
issn = {1361-8415},
doi = {https://doi.org/10.1016/j.media.2019.101552},
url = {https://www.sciencedirect.com/science/article/pii/S1361841518308430},
author = {Xin Yi and Ekta Walia and Paul Babyn}
}

@article{kazerouni2023diffusion,
title = {Diffusion models in medical imaging: A comprehensive survey},
journal = {Medical Image Analysis},
volume = {88},
pages = {102846},
year = {2023},
issn = {1361-8415},
doi = {https://doi.org/10.1016/j.media.2023.102846},
url = {https://www.sciencedirect.com/science/article/pii/S1361841523001068},
author = {Amirhossein Kazerouni and Ehsan Khodapanah Aghdam and Moein Heidari and Reza Azad and Mohsen Fayyaz and Ilker Hacihaliloglu and Dorit Merhof},
}

@article{schroeder1987coated,
author = {Kenneth W. Schroeder  and William J. Tremaine  and Duane M. Ilstrup },
title = {Coated Oral 5-Aminosalicylic Acid Therapy for Mildly to Moderately Active Ulcerative Colitis},
journal = {New England Journal of Medicine},
volume = {317},
number = {26},
pages = {1625-1629},
year = {1987},
doi = {10.1056/NEJM198712243172603},
URL = {https://www.nejm.org/doi/full/10.1056/NEJM198712243172603},
eprint = {https://www.nejm.org/doi/pdf/10.1056/NEJM198712243172603}
}

@inproceedings{mou2024t2i,
  title={T2I-Adapter: Learning Adapters to Dig out More Controllable Ability for Text-to-Image Diffusion Models},
  author={Chong Mou and Xintao Wang and Liangbin Xie and Jing Zhang and Zhongang Qi and Ying Shan and Xiaohu Qie},
  booktitle={AAAI Conference on Artificial Intelligence},
  year={2023},
  url={https://api.semanticscholar.org/CorpusID:256900833}
}

@inproceedings{Kynknniemi2019ImprovedPA,
  title={Improved Precision and Recall Metric for Assessing Generative Models},
  author={Tuomas Kynk{\"a}{\"a}nniemi and Tero Karras and Samuli Laine and Jaakko Lehtinen and Timo Aila},
  booktitle={Neural Information Processing Systems},
  year={2019},
  url={https://api.semanticscholar.org/CorpusID:118648975}
}

@article{ROUSSEEUW198753,
title = {Silhouettes: A graphical aid to the interpretation and validation of cluster analysis},
journal = {Journal of Computational and Applied Mathematics},
volume = {20},
pages = {53-65},
year = {1987},
issn = {0377-0427},
doi = {https://doi.org/10.1016/0377-0427(87)90125-7},
url = {https://www.sciencedirect.com/science/article/pii/0377042787901257},
author = {Peter J. Rousseeuw},
}
}

\end{document}